\documentclass{article}

\usepackage{arxiv}

\usepackage[utf8]{inputenc} 
\usepackage[T1]{fontenc}    
\usepackage{hyperref}       
\usepackage{url}            
\usepackage{booktabs}       
\usepackage{amsfonts}       
\usepackage{nicefrac}       
\usepackage{microtype}      
\usepackage{lipsum}
\usepackage{amsmath}

\title{Medical Concept Normalization in User Generated Texts by Learning Target Concept Embeddings}

\author{
  Katikapalli Subramanyam Kalyan \thanks{corresponding author} \\
  Department of Computer Applications\\
  NIT Trichy, India\\
  \texttt{kalyan.ks@yahoo.com} \\
   \And
 S.Sangeetha \\
  Department of Computer Applications\\
  NIT Trichy, India\\
  \texttt{sangeetha@nitt.edu} \\
}

\date{}
\begin{document}
\date{}
\maketitle

\begin{abstract}
Medical concept normalization helps in discovering standard concepts in free-form text i.e., maps health-related mentions to standard concepts in a vocabulary.  It is much beyond simple string matching and requires a deep semantic understanding of concept mentions. Recent research approach concept normalization as either text classification or text matching. The main drawback in existing a) text classification approaches is ignoring valuable target concepts information in learning input concept mention representation b) text matching approach is the need to separately generate target concept embeddings which is time and resource consuming. Our proposed model overcomes these drawbacks by jointly learning the representations of input concept mention and target concepts. First, it learns input concept mention representation using RoBERTa. Second, it finds cosine similarity between embeddings of input concept mention and all the target concepts. Here, embeddings of target concepts are randomly initialized and then updated during training. Finally, the target concept with maximum cosine similarity is assigned to the input concept mention. Our model surpass all the existing methods across three standard datasets by improving accuracy up to 2.31\%.
\end{abstract}


\section{Background}
Internet users use social media to voice their views and opinions. Medical social media is a part of social media in which the focus is limited to health and related issues \cite{pattisapu2017medical}. User generated texts in medical social media include tweets, blog posts, reviews on drugs, health related question and answers in discussion forums.  This rich source of data can be utilized in many health related applications to enhance the quality of services provided \cite{kalyan2020secnlp}. However, analysis of user generated texts is challenging due to noisy nature. User generated texts are written using informal language in descriptive words with irregular abbreviations, slang words and emoticons. 

Medical concept normalization aims at discovering medical concepts in free-form text. In this task, health related mentions are mapped to standard concepts in a clinical knowledge base. Common public express health related conditions in an informal way using layman terms while clinical knowledge base contain concepts expressed in scientific language. This variation (colloquial vs scientific) in the languages of common public and knowledge bases makes concept normalization an essential step in understanding user generated texts. This task is much beyond simple string matching as the same concept can be expressed in a descriptive way using colloquial words or in  multiple ways using aliases, acronyms, partial names and morphological variants. Further, noisy nature of user generated texts and the short length of health related mentions makes the task of concept normalization more challenging.

Research in medical concept normalization started with string matching techniques \cite{aronson2001effective, mccallum2005conditional,tsuruoka2007learning} followed by machine learning techniques \cite{leaman2013dnorm,leaman2014automated}. The inability of these methods to consider semantics into account shifted research towards deep learning methods with embeddings as input \cite{limsopatham2016normalising,lee2017medical,tutubalina2018medical,SUBRAMANYAM20201353}. For example, Lee et al.\cite{lee2017medical} and Tutubalina et al. \cite{tutubalina2018medical} experimented with RNN on the top of domain specific embeddings. Further, lack of large labeled datasets and necessity to train deep learning models like CNN or RNN from scratch (except embeddings) shifted research towards using pretrained language models like BERT  and RoBERTa \cite{miftahutdinov2019deep,kalyan2020bertmcn,pmlr-v116-pattisapu20a}. Miftahutdinov and Tutubalina \cite{miftahutdinov2019deep} experimented with BERT based fine-tuned models while Kalyan and Sangeetha \cite{kalyan2020bertmcn} provided a comprehensive evaluation of BERT based general and domain specific models. Recently, Pattisapu et al. \cite{pmlr-v116-pattisapu20a} work is based on RoBERTa \cite{liu2019roberta} and SNOMED-CT graph embeddings. In this work, concept normalization is viewed as text matching problem and appropriate standard concept is chosen based on cosine similarity between RoBERTa encoded concept mention and target concept embeddings. 

The main drawback in existing
\begin{itemize}
    \item text classification approaches \cite{tutubalina2018medical,miftahutdinov2019deep,kalyan2020bertmcn} is not exploiting the target concepts information in learning input concept mention representation. However, recent work in various natural language processing and computer vision tasks highlights the importance of exploiting  target label information in learning input representation. \cite{rodriguez2013label,akata2015label,wang2018joint,pappas2019gile,liu2020label}.
    \item text matching approach of \cite{pmlr-v116-pattisapu20a} is the need to generate target concept embeddings separately using text or graph embedding methods. This is time and resource consuming when different vocabularies are used for mapping in different data sets (e.g., SNOMED-CT  is used  in CADEC \cite{karimi2015cadec} and PsyTAR  \cite{zolnoori2019systematic} data sets, MedDRA is used in SMM4H2017 \cite{sarker2018data}). Moreover, the quality of generated concept embeddings depends on the comprehensiveness of vocabulary. For example, MedDRA is less fine grained compared to SNOMED-CT \cite{bodenreider2009using}. This requirement of comprehensive vocabulary limits the effectiveness of this approach. 
\end{itemize}

 Our model normalizes input concept mention by jointly learning the representations of input concept mention and target concepts. By learning the representations of target concepts along with input concept mention, our model a) exploits target concepts information unlike existing text classification approaches \cite{tutubalina2018medical,miftahutdinov2019deep,kalyan2020bertmcn} and b) eliminate the need to separately generate target concept embeddings unlike existing text matching approach \cite{pmlr-v116-pattisapu20a} which is time and resource consuming. Our model achieves the best results across three standard data sets surpassing all existing methods with an accuracy improvement of up to 2.31\%.

\section{Methodology}

\subsection{Model Description}

Our model normalizes concept mentions in two phases. First, it learns input concept mention representation using RoBERTa \cite{liu2019roberta}. Second, it finds cosine similarity between embeddings of input concept mention and all the target concepts. Here, embeddings of target concepts are randomly initialized and then updated during training. Finally, the target concept with maximum cosine similarity is assigned to the input concept mention.

Input concept mention is encoded into a fixed size vector $\boldsymbol{m} \in \mathbb{R}^d$ using RoBERTa. RoBERTa  is a contextualized embedding model pre-trained on 160 GB of text corpus using language modeling objective. It consists of an embedding layer followed by a sequence of transformer encoders \cite{liu2019roberta}. 
\begin{equation}
    m = RoBERTa(mention)
\end{equation}

Input concept mention vector $\boldsymbol{m}$ is transformed into cosine similarity vector $ \boldsymbol{q}$ by finding cosine similarity between m and randomly initialized embeddings $\{\boldsymbol{c_1},\boldsymbol{c_2},\boldsymbol{c_3},…\boldsymbol{c_N}\}$ of all target concepts $\{\boldsymbol{C_1},\boldsymbol{C_2},…\boldsymbol{C_N}\}$ where $\boldsymbol{c_i} \in \mathbb{R}^d$ and $N$ represents total number of unique concepts in the dataset. During training, the target concept embeddings and parameters of RoBERTa are updated. 
\begin{equation}
    \boldsymbol{q}=[q_i]_{i=1}^{N} \, where \, q_i=CS(\boldsymbol{m},\boldsymbol{c_i})
\end{equation}
\noindent Here $i=1,2,3,…N$ and $CS()$ represents cosine similarity defined as 
\begin{equation}
    CS(\boldsymbol{m},\boldsymbol{c}) = \frac{\boldsymbol{\mathit{m}}. \boldsymbol{\mathit{c}}} {\| \boldsymbol{m} \|.\| \boldsymbol{c} \|} 
    = \frac{ \sum_{i=1}^{N} m_i \times c_i }{ \sqrt{\sum_{i=1}^{N} (m_i)^2} \times \sqrt{\sum_{i=1}^{N} (c_i)^2}}
\end{equation} Cosine similarity vector $\boldsymbol{q}$ is normalized to $\boldsymbol{\hat{q}}$  using softmax function.
\begin{equation}
    \boldsymbol{\hat{q}} = Softmax(\boldsymbol{q})
\end{equation}

Finally, model is trained by minimizing the cross entropy loss between normalized cosine similarity vector $\hat{q}$ and one hot encoded ground truth vector $p$. Here $M$ represents number of training instances.
\begin{equation}
   \mathfrak{L} = - \sum_{i=1}^{M} {\sum_{j=1}^{N} {p_j^{i}} log(\hat{q}_j^{i})}  
\end{equation}

\subsection{Evaluation}
We evaluate our normalization system using accuracy metric, as in the previous works \cite{miftahutdinov2019deep, kalyan2020bertmcn, pmlr-v116-pattisapu20a}. Accuracy represents the percentage of correctly normalized mentions. In case of CADEC \cite{karimi2015cadec} and PsyTAR \cite{zolnoori2019systematic} datasets which are multi-fold, reported accuracy is average accuracy across folds. 

\section{Experimental Setup}
\subsection{Implementation Details}
Preprocessing of input concept mentions include a) removal of non-ASCII and special characters b) normalizing words with more than two consecutive repeating characters (e.g., sleeep $\to$ sleep) c) replacing English contraction and medical acronym words with corresponding full forms (e.g., can’t $\to$ cannot, bp $\to$ blood pressure). The list of medical acronyms is gathered from acronymslist.com and Wikipedia.  

We choose 10\% of training set for validation and find optimal hyperparameter values using random search. We use AdamW optimizer \cite{loshchilov2018decoupled} with a learning rate of 3e-5. The final results reported are based on the optimal hyperparameter settings. 

\subsection{Datasets}
\textbf{SMM4H2017} : This dataset is released for subtask3 of SMM4H 2017 \cite{sarker2018data}. It consists of ADR phrases extracted from twitter using drug names as keywords and then mapped to Preferred Terms (PTs) from MedDRA. In this, training set includes 6650 phrases assigned with 472 PTs and test set includes 2500 phrases assigned with 254 PTs. 

\noindent \textbf{CADEC}: CSIRO Adverse Drug Event Corpus (CADEC) includes user generated medical reviews related to Diclofenac and Lipitor \cite{karimi2015cadec}.  The manually identified health related mentions are mapped to target concepts in SNOMED-CT vocabulary. The dataset includes 6,754 mentions mapped one of  the 1029 SNOMED-CT codes. As the random folds of CADEC dataset created by Limsopatham and Collier \cite{limsopatham2016normalising} have significant overlap between train and test instances, Tutubalina et al. \cite{tutubalina2018medical} create custom folds of this dataset with minimum overlap. 

\noindent \textbf{PsyTAR}: Psychiatric Treatment Adverse Reactions (PsyTAR) corpus includes psychiatric drug reviews obtained from AskaPatient \cite{zolnoori2019systematic}. Zolnoori et al. \cite{zolnoori2019systematic} manually identify 6556 health related mentions and map them to one of 618 SNOMED-CT codes. Due to significant overlap between train and test sets of random folds released by Zoolnori et al. \cite{zolnoori2019systematic}, Miftahutdinov and Tutubalina \cite{miftahutdinov2019deep} create custom folds of this dataset with minimum overlap.

\section{Results}

\begin{table*}[t!]
\begin{center}
\begin{tabular}{|l|l|l|l|}
\hline \footnotesize  Method &  \footnotesize CADEC &  \footnotesize PsyTAR	&  \footnotesize SMM4H17 \\ \hline
\footnotesize Miftahutdinov and Tutubalina \cite{miftahutdinov2019deep} & \footnotesize	79.83 &\footnotesize 77.52 & \footnotesize	89.64 \\
\footnotesize Kalyan and Sangeetha \cite{kalyan2020bertmcn} &	\footnotesize 82.62 &\footnotesize	- & \footnotesize	- \\
\footnotesize Pattisapu et al. \cite{pmlr-v116-pattisapu20a} & \footnotesize	83.18 & \footnotesize	82.42 & \footnotesize	- \\ \hline
\footnotesize Roberta-base + concept embeddings\textsuperscript{$\bot$} & \footnotesize	82.60 &	 \footnotesize 81.90 &  \footnotesize 90.15 \\
\footnotesize Roberta-large + concept embeddings\textsuperscript{$\bot$} & \footnotesize \textbf{85.49} &	\footnotesize \textbf{83.68} &	\footnotesize \textbf{90.84}\\
\hline
 \end{tabular}
\end{center}
\caption{\label{results} Accuracy of existing methods and our proposed model across CADEC, PsyTAR and SMM4H2017 datasets. $\bot$ - concept embeddings are randomly initialized and then updated during training.} 
\end{table*}

Table \ref{results} provides a comparison of our model and the existing methods across three standard concept normalization datasets CADEC, PsyTAR and SMM4H2017. The first three rows represent existing systems and the next two rows represent our approach. Our model achieves the new state-of-the-art accuracy of 85.49\%, 83.68\% and 90.84\% across the three datasets. Our model outperforms existing methods with accuracy improvement of 2.31\%, 1.26\% and 1.2\% respectively. State-of-the-art results achieved by our model across three standard datasets illustrate that learning target concept representations along with input mention representations is simple and much effective compared to separately generating target concept representations using text or graph embedding methods and then using them. 

Robert-large   based model outperforms existing methods across all the three datasets but Robert-base based model outperforms existing methods only in case of SMM4H2017. This is because of relatively small sizes of CADEC and PsyTAR datasets compared to SMM4H2017. 

\section{Analysis}
Here, we discuss merits and demerits of our proposed method.
\subsection{ Merit Analysis}
We illustrate the effectiveness of our approach in the following two cases. 
\begin{itemize}
    \item In case I, existing methods map the concept mention `\textit{no concentration}' to a closely related target concept `\textit{Poor concentration (26329005)}' instead of the correct target concept `\textit{Unable to concentrate (60032008)}'. Similarly, `\textit{sleepy}' is mapped to `\textit{hypersomnia (77692006)}' instead of `\textit{drowsy (271782001)}'.
    \item In case II, `\textit{horrible pain}' is  mapped to abstract target concept `\textit{Pain (22253000)}' instead of fine-grained target concept `\textit{Severe pain (76948002)}'. Similarly, `\textit{fatigue in arms}' is mapped to `\textit{fatigue (84229001)}' instead of `\textit{muscle fatigue (80449002)}'.
\end{itemize}

In both the cases, existing methods are unable to exploit target concept information effectively and fail to assign the correct concept. However, our approach exploits target concept information by jointly learning representations of input concept mention and target concepts and hence assigns the concepts correctly. 
\subsection{Demerit Analysis}
 Our model aims to map health related mentions to standard concepts. We observe the predictions of our model and identify two groups of errors. 
 \begin{itemize}
     \item In case I, errors are related to insufficient number of training instances. For example, `\textit{hard to stay awake}' is assigned with more frequent concept `\textit{insomnia (193462001)}' instead of the ground truth concept `\textit{drowsy (271782001)}'. Similarly `\textit{muscle cramps in lower legs}' is assigned with `\textit{cramp in lower limb (449917004)}' instead of `\textit{cramp in lower leg (449918009)}'.
     \item In case II, errors are related to the inability in learning appropriate representations for domain specific rare words. For example, the mentions `\textit{pruritus}' and `\textit{hematuria}' are assigned to completely unrelated concepts `\textit{Tinnitus (60862001)}' and `\textit{diarrhea (62315008)}' respectively. 
 \end{itemize}
 
\section{Conclusion} 
In this work, we deal with medical concept normalization in user generated texts. Our model overcomes the drawbacks in existing text classification and text matching approaches by jointly learning the representations of input concept mention and target concepts. By learning target concept representations along with input concept mention representations, our approach a) exploits valuable target concept information unlike existing text classification approaches and b) eliminates the need to separately generate target concept embeddings unlike existing text matching approach. Our model surpass all the existing methods across three standard datasets by improving accuracy up to 2.31\%. In future, we would like to explore other possible options to include target concept information which may further improve the results.

\bibliographystyle{unsrt} 
\bibliography{references} 

\end{document}